\documentclass[runningheads]{llncs}
\usepackage[T1]{fontenc}
\usepackage{graphicx,verbatim}
\usepackage{multirow}
\usepackage{booktabs}
\usepackage{amsfonts}
\usepackage{amsmath} 
 \usepackage{subcaption}
\usepackage{xcolor}
\usepackage{ulem}

\begin{document}
\title{SAW: Toward a Surgical Action World Model via Controllable and Scalable Video Generation}
%
\author{Sampath Rapuri\inst{1,\star} \and
Lalithkumar Seenivasan\inst{1,\star} \and
Dominik Schneider\inst{1} \and
Roger Soberanis-Mukul\inst{1} \and
Yufan He\inst{2} \and
Hao Ding\inst{1} \and 
Jiru Xu\inst{1} \and
Chenhao Yu\inst{1} \and
Chenyan Jing\inst{1} \and
Pengfei Guo\inst{2} \and
Daguang Xu\inst{2} \and
Mathias Unberath\inst{1}}
\authorrunning{Rapuri et al.}
\institute{Johns Hopkins University, Baltimore MD, USA
\and
NVIDIA}


\maketitle              
\begin{abstract}
A surgical world model capable of generating realistic surgical action videos with precise control over tool-tissue interactions can address fundamental challenges in surgical AI and simulation -- from data scarcity and rare event synthesis to bridging the sim-to-real gap for surgical automation. However, current video generation methods, the very core of such surgical world models, require expensive annotations or complex structured intermediates as conditioning signals at inference, limiting their scalability. Other approaches exhibit limited temporal consistency across complex laparoscopic scenes and do not possess sufficient realism. We propose Surgical Action World (SAW) -- a step toward surgical action world modeling through video diffusion conditioned on four lightweight signals: language prompts encoding tool-action context, a reference surgical scene, tissue affordance mask, and 2D tool-tip trajectories. We design a conditional video diffusion approach that reformulates video-to-video diffusion into trajectory-conditioned surgical action synthesis. The backbone diffusion model is fine-tuned on a custom-curated dataset of $12,044$ laparoscopic clips with lightweight spatiotemporal conditioning signals, leveraging a depth consistency loss to enforce geometric plausibility without requiring depth at inference. SAW achieves state-of-the-art temporal consistency (CD-FVD: 199.19 vs. 546.82) and strong visual quality on held-out test data. Furthermore, we demonstrate its downstream utility for (a) surgical AI, where augmenting rare actions with SAW-generated videos improves action recognition (clipping F1-score: 20.93\% to 43.14\%; cutting: 0.00\% to 8.33\%) on real test data, and (b) surgical simulation, where rendering tool-tissue interaction videos from simulator-derived trajectory points toward a visually faithful simulation engine.


\end{abstract}

\section{Introduction}
A surgical world model that enables controllable, scalable video synthesis of surgical actions with realistic tool-tissue interaction has the potential to transform both surgical AI and the next generation of high-fidelity surgical simulators. For surgical AI, it offers a path to overcome the fundamental data scarcity that is a bottleneck for developing perception models, and enables targeted synthesis of rare but clinically critical events without costly setups or tissue use~\cite{wagner2023comparative,zia2025surgical,twinanda2016endonet,sharma2023rendezvous}. For simulation, such a world model can bridge the persistent sim-to-real gap by enabling visually realistic closed-loop environments where instrument motion drives controllable tool-tissue interactions. This supports both accelerated curation of paired kinematics-video data necessary for automating surgical actions and the development of digital twins for surgical safety evaluation through forward simulation of tool actions~\cite{tagliabue2020soft,scheikl2022sim,ha2018world,yang2023learning}. 
Realizing such a surgical world model hinges on advances in controllable and realistic video generation. While recent surgical video generation methods~\cite{biagini2025hierasurg,sivakumar2025sg2vid,chen2025surgsora} have demonstrated potential in producing visually plausible outputs, they remain limited in inference-time controllability and scalability: HieraSurg relies on expensive dense annotations such as per-frame segmentation masks~\cite{biagini2025hierasurg}; SG2VID depends on structured intermediates (spatio-temporal scene graphs) that are difficult to obtain or manipulate at inference~\cite{sivakumar2025sg2vid}; and even more flexible trajectory-conditioned approaches like SurgSora are constrained by limited inference windows ($W=21$) and struggle to maintain temporal consistency across complex laparoscopic dynamics~\cite{chen2025surgsora}.
%
%

To this end, we propose Surgical Action World (SAW), a step toward a surgical action world model. SAW enables controllable and scalable video diffusion conditioned on four lightweight signals: (i) structured language prompts encoding instrument and action context, (ii) a reference first frame anchoring scene appearance, (iii) tissue affordance masks specifying interaction regions, and (iv) 2D tool-tip trajectory across the surgical scene. \textbf{Our key contributions are}: (1) a curated dataset of 12,044 laparoscopic video clips with spatiotemporal annotations including video-level tool-action labels and tissue affordance along with frame-level tool-tip trajectories; (2) a video diffusion approach that reformulates video-to-video generation for controllable surgical action synthesis, with a novel depth consistency loss enforcing geometric plausibility without requiring depth at inference; and (3) demonstration of two downstream applications -- augmentation of rare surgical actions to improve action recognition on real test data, and rendering realistic tool-tissue interactions from simulator-derived tool trajectories to drive visually faithful surgical simulation.

\section{Surgical Action World (SAW) Model}
To advance surgical world modeling, we propose Surgical Action World (SAW), which reformulates video-to-video diffusion for controllable surgical action synthesis. By conditioning on instrument trajectories and tissue affordance alongside standard text and image inputs, SAW models tool-tissue interaction dynamics for scalable and realistic video generation (Fig.~\ref{fig:training_description}). The diffusion process is defined as $z_{i-1} = \mathcal{D}(z_i, t_i, \mathbf{z^a}, \mathbf{z^f}, \mathbf{z^p}, \mathbf{z^\gamma})$, where $z_i$ is the latent embedding from the previous denoising stage, $t_i$ the denoising iteration with $i = 1,\dots,N=50$, and $z_0 = x_0$ the clean video. Generation is conditioned on four lightweight signals: a language prompt $\mathbf{z^a}$ encoding tool-action context, a reference frame $\mathbf{z^f}$ anchoring scene appearance, an instrument trajectory $\mathbf{z^p}$ controlling tool-tip motion, and a tissue affordance map $\mathbf{z^\gamma}$ specifying interaction regions.

\subsubsection{2.1 Video Diffusion Model:}
\begin{figure}[!t]
    \centering
    \includegraphics[width=0.85\linewidth]{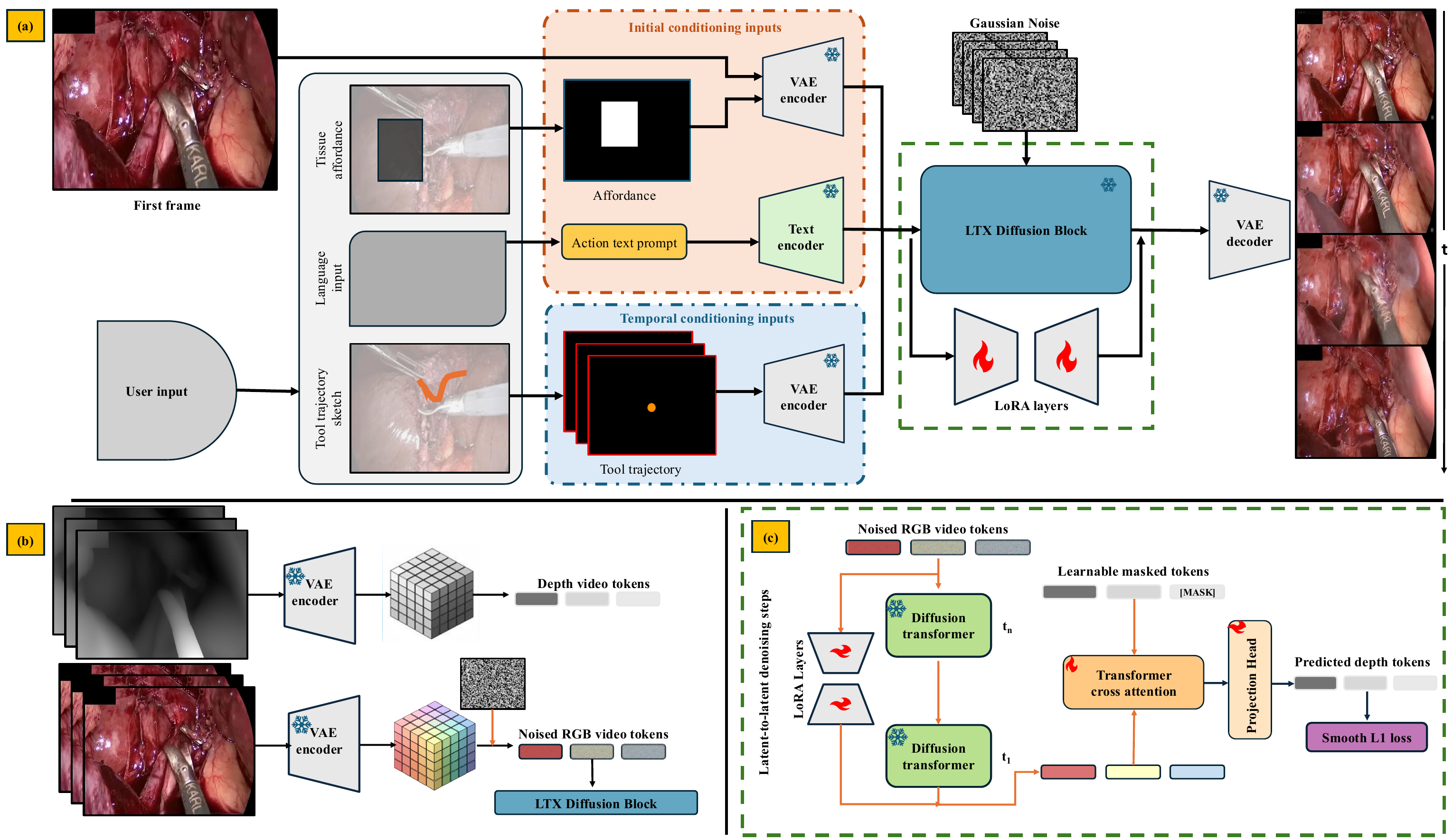}
    \caption{\textbf{a)} At inference, SAW leverages a user-inputted 2D tool trajectory, language prompt, tissue affordance, and a background surgical scene, enabling realistic and controllable video diffusion. \textbf{b)} During training, normalized depth and RGB videos are encoded using a frozen VAE, while the RGB video embedding is then flattened and noised into a sequence of noisy tokens. \textbf{c)} A depth consistency loss ($\mathcal{L}_{\text{DC}}$) is computed during training from the predicted masked depth video tokens solely using the denoised RGB video tokens.}
    \label{fig:training_description}
\end{figure}

We adopt LTX-Video~\cite{hacohen2024ltx} as the backbone for SAW, a transformer-based latent diffusion model that natively supports multi-modal conditioning (text, video, image). LTX performs diffusion in latent space using a variational autoencoder (VAE) for spatiotemporal downsampling, with denoising via a diffusion transformer trained using flow-matching~\cite{lipman2022flow}. We fine-tune LTX using In-Context Low Rank Adaptation (IC-LoRA) for surgical action synthesis with four conditioning embeddings: a reference frame $\mathbf{z^f}$, language prompt $\mathbf{z^a}$, tissue affordance map $\mathbf{z^\gamma}$, and temporal tool-tip trajectory $\mathbf{z^p}$, enabling the model to generate temporally coherent surgical action videos with realistic tool-tissue interactions (Fig.~\ref{fig:training_description}a).

\subsubsection{2.2 Lightweight Conditioning Signals:}
\uline{Language Prompting ($\mathbf{z^a}$):} 
To incorporate tool-action context, we condition on language prompts via LTX's text mechanism, following the template: ``A robotic da Vinci \{tool\}\ performs \{action\}\ during a cholecystectomy. The laparoscopic camera view shows the \{tool\} within the abdominal cavity. The \{tool\}\ moves precisely as it executes the \{action\} motion.''
\uline{Affordance Conditioning ($\mathbf{z^\gamma}$):}
To guide where tool-tissue interaction occurs, we condition the model on a 2D binary affordance map $I_a$ defined with respect to the reference frame $I_f$. The affordance embedding $\mathbf{z^\gamma}$ is extracted via a VAE encoder.
\uline{Tool Trajectory ($\mathbf{z^p}$):}
To precisely control instrument tip motion and tool-tissue interaction, we encode the tool trajectory as a temporal sequence of 2D maps $I_p^i \in \mathbb{R}^{h\times w \times 3}$, $i \in [0, k]$, where $k$ is the sequence length.
A trajectory point at time $i$ is represented as a fixed-radius circle centered at the tool-tip $(x,y)$ in $I_p^i$, with instrument class encoded through the R and G channels. The embedding $\mathbf{z^p}$ is obtained by passing the sequence through a VAE encoder.

\subsubsection{2.3 Depth Consistency Loss:}
In the presented formulation, all spatial conditioning signals are expressed as 2D inputs, with control in the Z-dimension learned implicitly by LTX during training. However, due to the presence of vital anatomy in the surgical scene, omitting depth guidance may result in inaccurate movements that violate critical safety constraints. This motivates the introduction of depth control during training, leading us to develop a novel loss for depth consistency ($\mathcal{L}_{\text{DC}}$). 
%
We first generate corresponding depth-mask videos for all RGB videos in our training dataset using Depth Anything V2 \cite{yang2024depth}.
After encoding these depth videos into the latent space (Fig.~\ref{fig:training_description}b), we introduced a cross-attention layer and a projection head that learns to reconstruct masked depth latents. This reconstruction uses the denoised RGB latent tokens created after a forward pass of the diffusion transformer. The depth consistency loss (Fig.~\ref{fig:training_description}c), $\mathcal{L}_{\text{DC}}$, is then computed using a Smooth $\ell_1$ loss, which follows an $\ell_2$ loss for small errors and an $\ell_1$ loss for large ones, providing robustness to outlier depth token predictions. Through this loss, we enforce geometric consistency in the Z-dimension without requiring explicit depth inputs at inference time.


\section{Experiment and Results}

\subsection{Experimental Setup}

\textbf{Dataset:} 
We custom-curate a dataset of $12,044$ video clips from 101 surgical videos sourced from 21 Youtube videos ($2,760$ video clips) and 80 videos ($12,878$ video clips) taken from four publicly available datasets: 10 videos from HeiChole ($1,794$ video clips)~\cite{wagner2023comparative}, 30 videos from Cholec80 ($6,042$ video clips)~\cite{twinanda2016endonet}, 32 videos from SurgVU ($4,278$ video clips)~\cite{zia2025surgical}, and 8 videos from CRCD ($764$ video clips)~\cite{oh2024expanded}. Each video clip was segmented to capture a \textit{single surgical action} and was manually annotated for \textit{video-level action} (clipping, grasping, cutting, or dissecting) and \textit{active tool usage} (grasper, hook, clipper, or scissors) as well as \textit{tissue affordance}, which indicates the tissue region undergoing tool-tissue interaction. Frame-wise annotations are provided for \textit{tool-tip positions} p = (x, y) in pixel coordinates. All video clips are preprocessed to remove black borders, letterboxing, and text overlay. We standardized all video clips to 81 frames at 25 fps, truncating longer videos and padding shorter ones by repeating the final frame. All videos were then processed at a resolution of $1024$x$576$. 
The dataset is partitioned based on the $101$ source videos, with the train set containing $11,502$ video clips (clipping = $443$, grasping = $1,964$, cutting = $273$, dissecting = $8,822$), and the test set comprising 542 video clips (clipping = $29$, grasping = $126$, cutting = $20$, dissecting = $367$). 

\noindent \textbf{Training and Inference} All our experiments are conducted using a single NVIDIA A100. We fine-tuned LTX using IC-LoRA for $7,500$ steps, with $\alpha =$ 128, lr = $2 \times 10^{-4}$, AdamW optimizer, and bfloat16 mixed precision training. To improve alignment with groundtruth videos and enhance generation diversity, we employ classifier-free guidance with first-frame conditioning in 20\% of cases. During inference, we use 50 denoising steps with a guidance scale = $3.5$. All generated videos are $81$ frames.

\noindent \textbf{Evaluation Metrics} We evaluate the visual quality of generated video using FVD~\cite{unterthiner2018towards}, CD-FVD~\cite{ge2024content}, SSIM~\cite{wang2004image}, PSNR, and LPIPS~\cite{zhang2018unreasonable} metrics. FVD assesses visual realism at the video level. 
\uline{CD-FVD is sensitive to temporal inconsistencies, making it relevant to evaluate the temporal consistency of tool-tissue interactions.}
We further report SSIM, PSNR, and LPIPS to quantify frame-wise structural and perceptual image quality.



\subsection{Results and Ablation Studies}

\begin{table*}[!b]
\centering
\caption{Quantitative evaluation of our model performance in video generation against state-of-the-art models on a held-out test set. WAN~\cite{wang2025wan}  and LTX$_b$~\cite{hacohen2024ltx} are used off-the-shelf whereas SurgSora~\cite{chen2025surgsora} is fine-tuned on our training dataset.}
\scalebox{0.85}{
\begin{tabular}{lccccccc}
\toprule
\textbf{Model} & \textbf{FVD} $\downarrow$ & \textbf{CD-FVD} $\downarrow$ & \textbf{SSIM} $\uparrow$ & \textbf{PSNR} $\uparrow$ & \textbf{LPIPS} $\downarrow$ \\
\midrule
WAN~\cite{wang2025wan}        & 439.60              & 429.67          & 0.5750           & 15.82     & 0.4480 \\  
LTX$_b$~\cite{hacohen2024ltx}    & 319.37              & 504.31          & 0.5630           & 15.07           & 0.4920           \\
SurgSora~\cite{chen2025surgsora}   & 541.61          & 546.82          & 0.4163          & 17.10          & \textbf{0.3825} \\
\textbf{Ours (SAW)}  & \textbf{224.28} & \textbf{199.19} & \textbf{0.5948} & \textbf{17.36} & 0.4100          \\
\bottomrule
\end{tabular}}
\label{tab:quant_results}
\end{table*}

\begin{figure}[!t]
    \centering
    \includegraphics[width=\linewidth]{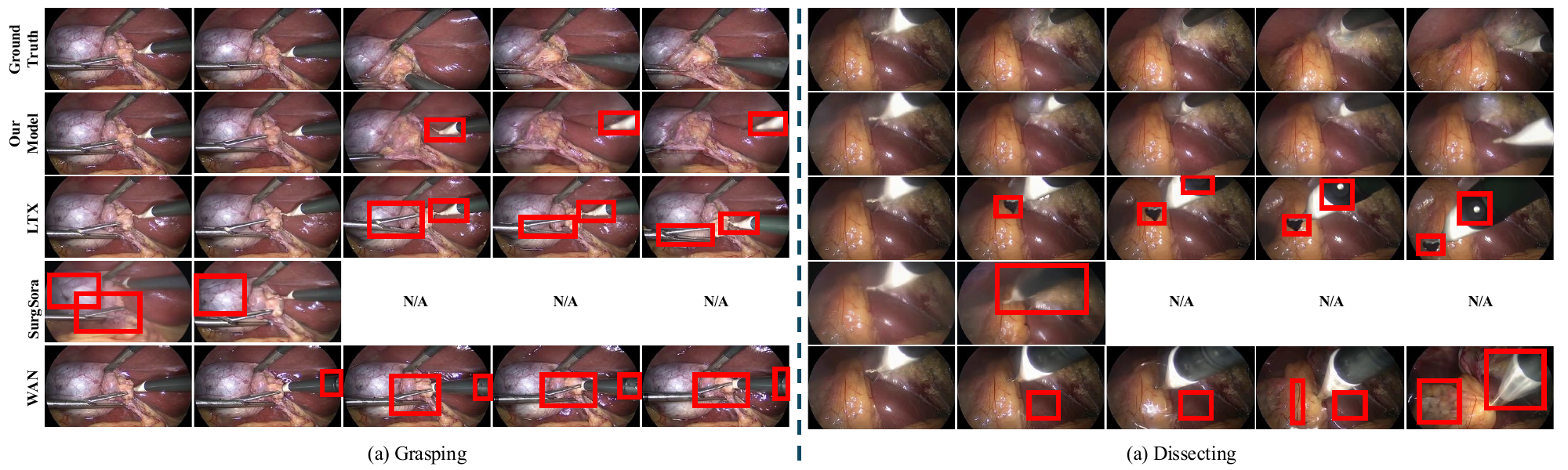}
    \caption{Qualitative examples of generated surgical videos for \textbf{(a)} grasping and \textbf{(b)} dissecting actions. Red overlays indicate where a generated video lacks realism or becomes visually distorted in comparison to the ground truth video at the corresponding frame.}
    \label{fig:qualitative_examples}
\end{figure}

\textbf{Baseline comparison:} Table~\ref{tab:quant_results} highlights the performance of our model compared to existing generative models. 
SAW achieves the lowest FVD at $224.28$. 
Moreover, it excels at generating temporally consistent videos, demonstrated through a CD-FVD of 199.19, outperforming the $546.82$ of the also trajectory-conditioned surgical generative model SurgSora.
For non-trajectory-conditioned models, the closest CD-FVD belongs to WAN with $429.67$, suggesting the benefit of incorporating a large pretraining dataset for video generation.
On frame-level content and structural metrics, SAW also outperforms all baselines with an SSIM of 0.5948 and PSNR of 17.36. SAW also demonstrated competitive 
frame-wise perceptual similarity with a LPIPS of 0.41. 
Fig.~\ref{fig:qualitative_examples} illustrates the quality of the generated videos across two separate actions. Even in the presence of endoscopic artifacts, our model can produce a coherent video that reflects the underlying action, demonstrating its controllability, robustness, and realism.

\begin{table*}[!t]
\centering
\caption{Ablation study across conditioning modalities and loss components. Each row removes a single conditioning signal or loss term from the full model. Best results per metric are in \textbf{bold}.}
\label{tab:ablation}
\scalebox{0.85}{
\begin{tabular}{l cccccc}
\toprule
\textbf{Ablated Component} &
\textbf{FVD} $\downarrow$ & \textbf{CD-FVD} $\downarrow$ & \textbf{LPIPS} $\downarrow$ & \textbf{SSIM} $\uparrow$ & \textbf{PSNR} $\uparrow$ \\
\midrule
SAW  & 224.28 & \textbf{199.19} & 0.4100 &  0.5948 & 17.36 \\
\midrule
\multicolumn{6}{c}{\textit{Ablation on Conditioning signals}} \\
\midrule
w/o Affordance       & 223.78          & 200.10 & 0.4101         & 0.5945          & 17.33 \\
w/o First Frame      & 1096.21         & 338.75          & 0.6217          & 0.4848          & 12.50 \\
w/o Language         & 219.23 & 204.89          & 0.4052 & \textbf{0.5977} & \textbf{17.47} \\
w/o Trajectory       & 278.26          & 344.79          & 0.4325          & 0.5818          & 16.45 \\
\midrule
\multicolumn{6}{c}{\textit{Model trained without depth consistency loss}} \\
\midrule
w/o $\mathcal{L}_{\text{DC}}$ & \textbf{188.34}             & 207.59          &  \textbf{0.4017}              & 0.5922              & 17.37 \\
\bottomrule
\end{tabular}}
\end{table*}

\noindent \textbf{Ablation Studies:} 
We assess the impact of our design choices and conditioning strategies in Table~\ref{tab:ablation}. 
Removing trajectory conditioning significantly impacts performance, as indicated by a consistent decrease in all the evaluation metrics.
Removing first-frame conditioning reduces visual quality, as indicated by drop performance in FVD, LPIPS, and SSIM. Furthermore, temporal consistency is affected as reflected by an increased CD-FVD to $338.75$.
Disabling the depth consistency loss, $\mathcal{L}_{\text{DC}}$, primary increases CD-FVD, suggesting a reduction in temporal consistency of tool and tissue movements. Overall, SAW with all design choices is the most balanced model across all evaluation metrics. 

\section{Downstream Applications}
We also demonstrate two downstream applications of our SAW model: (a) synthesizing rare surgical action videos to address class imbalance and enhance model performance on a real test set, and (b) exploring its potential to serve as a simulation engine for generating realistic tool–tissue interactions from surgical scenes and simulator-derived instrument trajectories.

\subsubsection{4.1 Synthetic Video Generation for Action Recognition:} 
\begin{figure} [!t]
    \centering
    \includegraphics[width=1\linewidth]{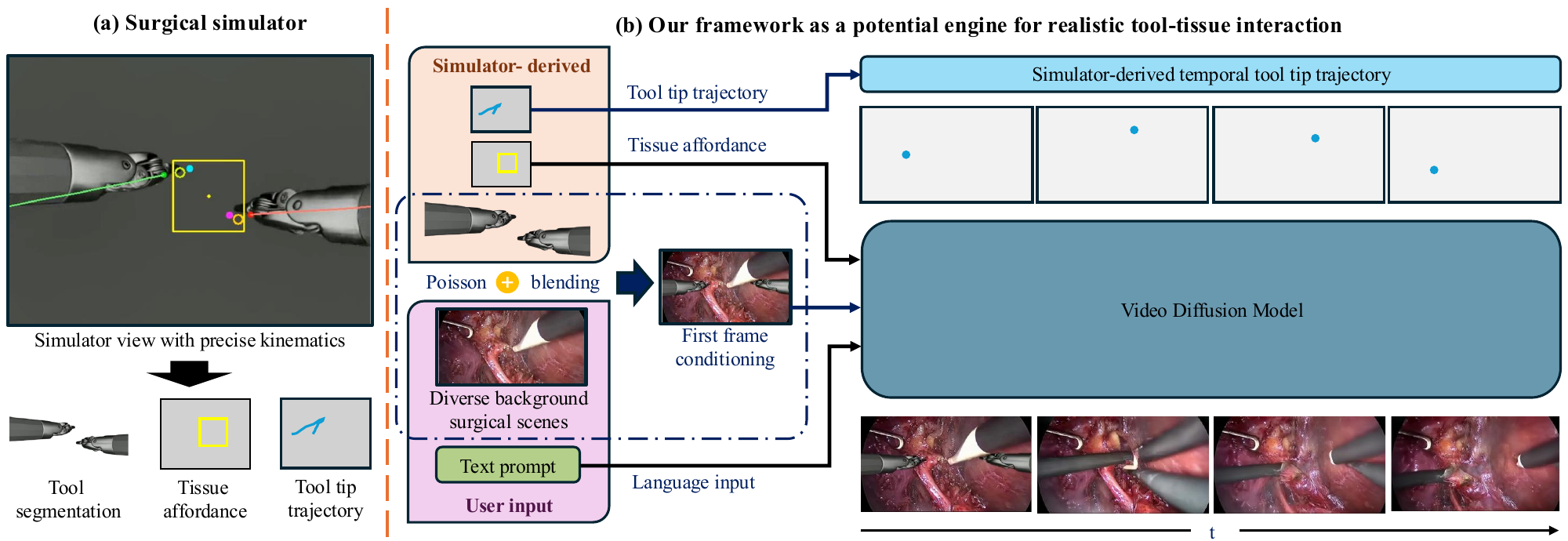}
    \caption{\textbf{a)} Simulator-derived tool segmentations, tissue affordance, tool tip trajectory. \textbf{b)} Simulator-derived tools are overlayed on a background surgical scene and used as conditional inputs to our video diffusion model, allowing us to model both tool kinematics and background tissue deformation.}
    \label{fig:simulator_generation_2}
\end{figure}

With action recognition being a major research topic in surgical AI~\cite{sharma2023rendezvous,liao2025artificial}, we design and conduct an experiment to demonstrate our framework's potential in augmenting rare action videos to improve model performance. Our train set exhibits a long-tail distribution, with cutting (273 clips) and clipping (443 clips) actions significantly underrepresented compared to dissecting ($8,822$ clips) and grasping ($1,964$ clips).  
\begin{table}[!t]
\centering
\caption{Per-class action recognition F1 score before and after video augmentation. Best results per action label are in \textbf{bold}.}
\label{tab:action_recognition}
\scalebox{0.85}{
\begin{tabular}{l|cc|cc}
\toprule
\multirow{2}{*}{\textbf{Actions}} & \multicolumn{2}{c|}{\textbf{Spatiotemporal CNN}~\cite{tran2018closer}} & \multicolumn{2}{c}{\textbf{ViT}~\cite{dosovitskiy2020image}} \\
                         & \textbf{w/o Aug} & \textbf{with Aug} & \textbf{w/o Aug} & \textbf{with Aug} \\
\midrule
Clipping   & 20.93          & \textbf{43.14} & \textbf{84.62} & 83.64 \\
Cutting    & 0.00           & \textbf{8.33}  & 30.77          & \textbf{47.06} \\
Grasping   & \textbf{58.52} & 55.87          & \textbf{74.19} & 73.80 \\
Dissection & 83.87          & \textbf{84.25} & 86.27          & \textbf{88.26} \\
\bottomrule
\end{tabular}}
\end{table}
\uline{Generating synthetic videos of rare events:}
Leveraging our controllable video generation framework, we generated synthetic videos of rare actions (cutting and clipping) to augment the training set. As Fig.~\ref{fig:simulator_generation_2}b describes, we overlay segmented surgical tools using SAM3 \cite{carion2025sam} onto diverse background surgical scenes, and leverage the tissue affordance and tool trajectory extracted from the source background video—substituted with the target tool class—to condition generation of the augmented video. Superimposed surgical tools, which are acquired from existing training video clips, are blended into the background scene using Poisson image blending \cite{perez2023poisson} to increase realism of the starting scene used to guide the diffusion process. While background scenes were diverse, we ensured relevance to each action -- for example, clipping actions were paired with scenes containing blood vessels, increasing the diversity of tool–anatomy pairings. In total, we generated 287 synthetic videos: 110 clipping and 177 cutting.
\uline{Experimental setup:}
We train two AI models -- a spatiotemporal CNN~\cite{tran2018closer} and a vision transformer (ViT) model~\cite{dosovitskiy2020image} -- for video action recognition. Consistent with state-of-the-art approaches in surgical action prediction~\cite{sharma2023rendezvous}, both models take as input 16-frame segments from each video to predict the action. To evaluate our framework's utility in enhancing model performance through data augmentation, we train each model under two settings: (i) using video clips only from the original training set, and (ii) augmenting the training set with synthetically generated video clips of rare actions.
\uline{Results and Discussion:} Table~\ref{tab:action_recognition} summarizes the performance of both models trained with and without synthetic rare action videos and evaluated on a held-out real test set. For both models, we observe that augmentation with synthetic videos can improve recognition of underrepresented actions (clipping and cutting) on real test videos, validating our framework's utility in generating realistic and effective training data.

\subsubsection{4.2 Realistic Tool Tissue Interaction Engine for Surgical Simulation:} 
While physics-based surgical simulators have made significant progress, they struggle to accurately model complex tissue-tool interactions and tissue deformation in real time~\cite{zhang2017deformable}. Towards addressing this challenge, we explore our controllable surgical action video generation framework as a potential engine to model realistic tool-tissue interaction and tissue deformation in near real-time. 
\uline{Qualitative Experimental Setup}: As shown in Fig.~\ref{fig:simulator_generation_2}, we build a surgical simulator using Isaac lab~\cite{mittal2025isaac}, encompassing surgical tools in an empty background. 
Given a set of simulator-derived data (instrument segmentation, their corresponding tool tip trajectories, and tissue affordance), we demonstrate our framework's flexibility and controllability  in generating surgical action videos -- modeling tool-tissue interaction with realistic tissue deformations -- with any given background surgical scene. However, this represents an initial proof-of-concept. Future work will address enforcing instrument joint kinematics, supporting more instruments and scenes, and achieving real-time inference.

\section{Conclusion}
We present Surgical Action World (SAW) -- a first step toward surgical action world modeling through controllable and scalable video diffusion. By conditioning on four lightweight signals (language prompts, reference frame, tissue affordance mask, and tool-tip trajectories), SAW achieves state-of-the-art temporal consistency and visual quality while enabling scalable inference without expensive annotations. A depth consistency loss further encourages geometric coherence in tool motion, promoting plausible tool-tissue interactions that extend beyond purely 2D trajectory conditioning. We demonstrate the downstream utility of SAW for surgical AI through rare action augmentation that improves action recognition on real data, and for simulation through rendering tool-tissue interactions from simulator-derived trajectories. Future work will focus on improving the integration of affordance and language cues for richer scene understanding, extending to longer video generation, incorporating additional instruments and anatomical scenes, and achieving real-time inference for closed-loop simulation.


\bibliographystyle{splncs04}
\bibliography{references}

@article{ha2018world,
  title={World models},
  author={Ha, David and Schmidhuber, J{\"u}rgen},
  journal={arXiv preprint arXiv:1803.10122},
  volume={2},
  number={3},
  pages={440},
  year={2018}
}

@article{yang2023learning,
  title={Learning interactive real-world simulators},
  author={Yang, Mengjiao and Du, Yilun and Ghasemipour, Kamyar and Tompson, Jonathan and Schuurmans, Dale and Abbeel, Pieter},
  journal={arXiv preprint arXiv:2310.06114},
  volume={1},
  number={2},
  pages={6},
  year={2023}
}

@article{wagner2023comparative,
  title={Comparative validation of machine learning algorithms for surgical workflow and skill analysis with the HeiChole benchmark},
  author={Wagner, Martin and M{\"u}ller-Stich, Beat-Peter and Kisilenko, Anna and Tran, Duc and Heger, Patrick and M{\"u}ndermann, Lars and Lubotsky, David M and M{\"u}ller, Benjamin and Davitashvili, Tornike and Capek, Manuela and others},
  journal={Medical image analysis},
  volume={86},
  pages={102770},
  year={2023},
  publisher={Elsevier}
}

@article{dosovitskiy2020image,
  title={An image is worth 16x16 words: Transformers for image recognition at scale},
  author={Dosovitskiy, Alexey and Beyer, Lucas and Kolesnikov, Alexander and Weissenborn, Dirk and Zhai, Xiaohua and Unterthiner, Thomas and Dehghani, Mostafa and Minderer, Matthias and Heigold, Georg and Gelly, Sylvain and others},
  journal={arXiv preprint arXiv:2010.11929},
  year={2020}
}

@inproceedings{tran2018closer,
  title={A closer look at spatiotemporal convolutions for action recognition},
  author={Tran, Du and Wang, Heng and Torresani, Lorenzo and Ray, Jamie and LeCun, Yann and Paluri, Manohar},
  booktitle={Proceedings of the IEEE conference on Computer Vision and Pattern Recognition},
  pages={6450--6459},
  year={2018}
}

@article{zhang2017deformable,
  title={Deformable models for surgical simulation: a survey},
  author={Zhang, Jinao and Zhong, Yongmin and Gu, Chengfan},
  journal={IEEE reviews in biomedical engineering},
  volume={11},
  pages={143--164},
  year={2017},
  publisher={IEEE}
}

@article{twinanda2016endonet,
  title={Endonet: a deep architecture for recognition tasks on laparoscopic videos},
  author={Twinanda, Andru P and Shehata, Sherif and Mutter, Didier and Marescaux, Jacques and De Mathelin, Michel and Padoy, Nicolas},
  journal={IEEE transactions on medical imaging},
  volume={36},
  number={1},
  pages={86--97},
  year={2016},
  publisher={IEEE}
}

@article{zia2025surgical,
  title={Surgical visual understanding (surgvu) dataset},
  author={Zia, Aneeq and Berniker, Max and Nespolo, Rogerio and Perreault, Conor and Wang, Ziheng and Mueller, Benjamin and Schmidt, Ryan and Bhattacharyya, Kiran and Liu, Xi and Jarc, Anthony},
  journal={arXiv preprint arXiv:2501.09209},
  year={2025}
}

@article{oh2024expanded,
  title={Expanded Comprehensive Robotic Cholecystectomy Dataset (CRCD)},
  author={Oh, Ki-Hwan and Borgioli, Leonardo and Mangano, Alberto and Valle, Valentina and Di Pangrazio, Marco and Toti, Francesco and Pozza, Gioia and Ambrosini, Luciano and Ducas, Alvaro and {\v{Z}}efran, Milo{\v{s}} and others},
  journal={arXiv preprint arXiv:2412.12238},
  year={2024}
}

@inproceedings{tagliabue2020soft,
  title={Soft tissue simulation environment to learn manipulation tasks in autonomous robotic surgery},
  author={Tagliabue, Eleonora and Pore, Ameya and Dall’Alba, Diego and Magnabosco, Enrico and Piccinelli, Marco and Fiorini, Paolo},
  booktitle={2020 IEEE/RSJ International Conference on Intelligent Robots and Systems (IROS)},
  pages={3261--3266},
  year={2020},
  organization={IEEE}
}

@article{scheikl2022sim,
  title={Sim-to-real transfer for visual reinforcement learning of deformable object manipulation for robot-assisted surgery},
  author={Scheikl, Paul Maria and Tagliabue, Eleonora and Gyenes, Bal{\'a}zs and Wagner, Martin and Dall'Alba, Diego and Fiorini, Paolo and Mathis-Ullrich, Franziska},
  journal={IEEE Robotics and Automation Letters},
  volume={8},
  number={2},
  pages={560--567},
  year={2022},
  publisher={IEEE}
}

@inproceedings{biagini2025hierasurg,
  title={Hierasurg: Hierarchy-aware diffusion model for surgical video generation},
  author={Biagini, Diego and Navab, Nassir and Farshad, Azade},
  booktitle={International Conference on Medical Image Computing and Computer-Assisted Intervention},
  pages={310--319},
  year={2025},
  organization={Springer}
}

@inproceedings{chen2025surgsora,
  title={Surgsora: Object-aware diffusion model for controllable surgical video generation},
  author={Chen, Tong and Yang, Shuya and Wang, Junyi and Bai, Long and Ren, Hongliang and Zhou, Luping},
  booktitle={International Conference on Medical Image Computing and Computer-Assisted Intervention},
  pages={521--531},
  year={2025},
  organization={Springer}
}

@inproceedings{sivakumar2025sg2vid,
  title={Sg2vid: Scene graphs enable fine-grained control for video synthesis},
  author={Sivakumar, Ssharvien Kumar and Frisch, Yannik and Ghazaei, Ghazal and Mukhopadhyay, Anirban},
  booktitle={International Conference on Medical Image Computing and Computer-Assisted Intervention},
  pages={511--521},
  year={2025},
  organization={Springer}
}

@incollection{perez2023poisson,
  title={Poisson image editing},
  author={P{\'e}rez, Patrick and Gangnet, Michel and Blake, Andrew},
  booktitle={Seminal Graphics Papers: Pushing the Boundaries, Volume 2},
  pages={577--582},
  year={2023}
}

@article{carion2025sam,
  title={Sam 3: Segment anything with concepts},
  author={Carion, Nicolas and Gustafson, Laura and Hu, Yuan-Ting and Debnath, Shoubhik and Hu, Ronghang and Suris, Didac and Ryali, Chaitanya and Alwala, Kalyan Vasudev and Khedr, Haitham and Huang, Andrew and others},
  journal={arXiv preprint arXiv:2511.16719},
  year={2025}
}

@article{mittal2025isaac,
  title={Isaac lab: A gpu-accelerated simulation framework for multi-modal robot learning},
  author={Mittal, Mayank and Roth, Pascal and Tigue, James and Richard, Antoine and Zhang, Octi and Du, Peter and Serrano-Munoz, Antonio and Yao, Xinjie and Zurbr{\"u}gg, Ren{\'e} and Rudin, Nikita and others},
  journal={arXiv preprint arXiv:2511.04831},
  year={2025}
}

@article{liao2025artificial,
  title={Artificial intelligence-assisted phase recognition and skill assessment in laparoscopic surgery: a systematic review},
  author={Liao, Wenqiang and Zhu, Ying and Zhang, Hanwei and Wang, Dan and Zhang, Lijun and Chen, Tianxiang and Zhou, Ru and Ye, Zi},
  journal={Frontiers in Surgery},
  volume={12},
  pages={1551838},
  year={2025},
  publisher={Frontiers Media SA}
}

@article{hacohen2024ltx,
  title={Ltx-video: Realtime video latent diffusion},
  author={HaCohen, Yoav and Chiprut, Nisan and Brazowski, Benny and Shalem, Daniel and Moshe, Dudu and Richardson, Eitan and Levin, Eran and Shiran, Guy and Zabari, Nir and Gordon, Ori and others},
  journal={arXiv preprint arXiv:2501.00103},
  year={2024}
}

@article{unterthiner2018towards,
  title={Towards accurate generative models of video: A new metric \& challenges},
  author={Unterthiner, Thomas and Van Steenkiste, Sjoerd and Kurach, Karol and Marinier, Raphael and Michalski, Marcin and Gelly, Sylvain},
  journal={arXiv preprint arXiv:1812.01717},
  year={2018}
}

@article{lipman2022flow,
  title={Flow matching for generative modeling},
  author={Lipman, Yaron and Chen, Ricky TQ and Ben-Hamu, Heli and Nickel, Maximilian and Le, Matt},
  journal={arXiv preprint arXiv:2210.02747},
  year={2022}
}

@article{yang2024depth,
  title={Depth anything v2},
  author={Yang, Lihe and Kang, Bingyi and Huang, Zilong and Zhao, Zhen and Xu, Xiaogang and Feng, Jiashi and Zhao, Hengshuang},
  journal={Advances in Neural Information Processing Systems},
  volume={37},
  pages={21875--21911},
  year={2024}
}

@inproceedings{zhang2018unreasonable,
  title={The unreasonable effectiveness of deep features as a perceptual metric},
  author={Zhang, Richard and Isola, Phillip and Efros, Alexei A and Shechtman, Eli and Wang, Oliver},
  booktitle={Proceedings of the IEEE conference on computer vision and pattern recognition},
  pages={586--595},
  year={2018}
}

@article{wang2004image,
  title={Image quality assessment: from error visibility to structural similarity},
  author={Wang, Zhou and Bovik, Alan C and Sheikh, Hamid R and Simoncelli, Eero P},
  journal={IEEE transactions on image processing},
  volume={13},
  number={4},
  pages={600--612},
  year={2004},
  publisher={IEEE}
}

@inproceedings{ge2024content,
  title={On the content bias in fr{\'e}chet video distance},
  author={Ge, Songwei and Mahapatra, Aniruddha and Parmar, Gaurav and Zhu, Jun-Yan and Huang, Jia-Bin},
  booktitle={Proceedings of the IEEE/CVF conference on computer vision and pattern recognition},
  pages={7277--7288},
  year={2024}
}

@article{wang2025wan,
  title={Wan: Open and advanced large-scale video generative models},
  author={Wang, Ang and Ai, Baole and Wen, Bin and Mao, Chaojie and Xie, Chen-Wei and Chen, Di and Yu, Feiwu and Zhao, Haiming and Yang, Jianxiao and Zeng, Jianyuan and others},
  journal={arXiv preprint arXiv:2503.20314},
  volume={3},
  number={4},
  pages={6},
  year={2025}
}

@article{sharma2023rendezvous,
  title={Rendezvous in time: an attention-based temporal fusion approach for surgical triplet recognition},
  author={Sharma, Saurav and Nwoye, Chinedu Innocent and Mutter, Didier and Padoy, Nicolas},
  journal={International journal of computer assisted radiology and surgery},
  volume={18},
  number={6},
  pages={1053--1059},
  year={2023},
  publisher={Springer}
}

\end{document}